\documentclass[runningheads]{llncs}
\usepackage[T1]{fontenc}
\usepackage{graphicx}
\usepackage{hyperref}
\makeatletter
\newcommand{\printfnsymbol}[1]{%
  \textsuperscript{\@fnsymbol{#1}}}   % 1 → *, 2 → †, …
\makeatother

\usepackage{listings}
\usepackage{xcolor}
\usepackage{comment}
\lstdefinelanguage{Turtle}{
  morekeywords={a},
  sensitive=true,
  morecomment=[l]{\#},
  morestring=[b]",
}
\let\oldtexttt\texttt
\renewcommand{\texttt}[1]{{\small\oldtexttt{#1}}}

\begin{document}
%
% \title{An ontology for cross-sectoral and public health crisis management}
\title{An ontology for cross-sectoral crisis management: core and public health modules}
\titlerunning{An ontology for cross-sectoral crisis management}
\author{
Aldo Gangemi\thanks{These authors contributed equally to this work.}\inst{1,3}\orcidID{0000-0001-5568-2684} \and
Rita T. Sousa\printfnsymbol{1}\inst{2}\orcidID{0000-0002-7241-8970} \and
Luigi Asprino\inst{3}\orcidID{0000-0003-1907-0677} \and
Giorgia Lodi\inst{4}\orcidID{0009-0000-9032-4831} \and
Andrea G. Nuzzolese\inst{4}\orcidID{0000-0003-2928-9496} \and
Valentina Presutti\inst{1}\orcidID{0000-0002-9380-5160} \and
Johannes Gysen\inst{5}\orcidID{0009-0002-8657-3725} \and 
Diana F. Sousa\inst{6}\orcidID{0000-0003-0597-9273} \and 
Luigi Spagnolo \inst{6}\orcidID{0009-0008-0179-7468}}
% %
\authorrunning{A. Gangemi and R.T. Sousa et al.}
\institute{University of Bologna, Bologna, Italy %\\
\email{\{aldo.gangemi,  valentina.presutti\}@unibo.it} \and
Data and Web Science Group, University of Mannheim, Mannheim, Germany %\\
\email{rita.sousa@uni-mannheim.de} \and
San Raffaele University of Rome, Rome, Italy 
\email{luigi.asprino@uniroma5.it}
\and
Institute for Cognitive Sciences and Technology, Italian National Research Council, Bologna, Italy %\\
\email{\{giorgia.lodi, andrea.nuzzolese\}@istc.cnr.it} \and
European Commission, Health Emergency Preparedness and Response Authority, Brussels, Belgium %\\
\email{han.gysen@ec.europa.eu} \and
European Commission, Joint Research Centre, Ispra, Italy %\\
\email{\{diana.francisco-de-sousa, luigi.spagnolo\}@ec.europa.eu}}
\maketitle              % typeset the header of the contribution
\begin{abstract}
% The abstract should briefly summarize the contents of the paper in

%The COVID‑19 pandemic, climate‑driven extreme events, the war in Ukraine, and rising geopolitical tensions illustrate the growing complexity of crises that span health, environment, security, and socioeconomic domains.
%Existing knowledge systems remain sector‑specific and ambiguous, thus hampering the whole‑of‑government coordination required by policy initiatives such as the European Preparedness Union Strategy. 

%The COVID-19 pandemic, climate disasters, the Ukraine war, and geopolitical tensions show how crises now cut across health, environment, security, and socioeconomic domains. Yet existing knowledge systems remain siloed and ambiguous undermining the cross-sector coordination demanded by initiatives like the European Preparedness Union Strategy. 
This paper presents the European Crisis Management Ontology (ECMO), a modular OWL-based ontology intended as a cross-sectoral reference for disaster risk reduction and response. 
ECMO is designed to be organised as a network of ontological modules. Among the modules, ECMO-CORE
%, the foundational module, 
captures fundamental crisis management concepts such as hazard, event, exposure, impact, and response measure and uses ontology design patterns and the OWL2 punning technique to resolve ambiguities between hazard types and event manifestations. In addition, domain-specific modules are defined as in the case of 
%The modular design enablesIn addiit domain-specific extensions. Specifically, we present an extension for 
the public health module aligned with SNOMED CT and ICD-11. To demonstrate the resource's utility, we used ECMO to represent the data of the Epidemic Intelligence from Open Sources system of the Joint Research Centre to generate an end-to-end pipeline that populates an ECMO-compliant knowledge graph from unstructured epidemiological news. Initial results demonstrate that ECMO provides the formal guardrails necessary for consistent and unified knowledge representation and integration. 
The ontology is publicly available at \url{https://doi.org/10.5281/zenodo.20070268} and is released under the Creative Commons Attribution 4.0 International (CC BY 4.0) license.

% This paper proposes the European Crisis Management Ontology (ECMO), a modular OWL‑based framework that integrates and aligns heterogeneous terminologies (e.g., UNDRR, SNOMED‑CT, ICD‑11) into a unified reference model. 
% ECMO captures core concepts—hazard, hazardous event, exposure, impact, and measures—while providing domain‑specific extensions, demonstrated here for public‑health emergencies. 
%We describe a hybrid top‑down/bottom‑up methodology, semi‑automatic ontology matching and expert curation to generate validated mappings. 
% Initial results show improved semantic interoperability and reduced ambiguity, potentially enabling more consistent knowledge extraction, reasoning, and decision support in high‑stakes crisis management. Future work will extend ECMO to additional sectors and embed it in operational knowledge‑graph pipelines.

\end{abstract}

% ---- Introduction ----
\section{Introduction}

The COVID-19 pandemic, increasingly frequent extreme weather events due to the climate crisis, the conflicts in Ukraine and Middle East are all recent and ongoing complex crises that affect multiple policy sectors. In such a challenging context, the need for a cross-sectoral and whole-of-government approach to preparedness and crisis management is emerging, advocating for more coordinated and effective policy interventions~\cite{niinisto2024safer}. The European Preparedness Union Strategy~\cite{ec2025joint} is one outstanding example of such a trend. 
In the scope of this policy strategy, the European Crisis Management Platform (ECMP) is an initiative led by the European Commission’s Secretariat General to enable interoperability and harmonisation across existing early warning and crisis management systems. The main aim is to ensure cross-sectoral information exchange and coordination during emergencies.

% To achieve a better cross-sectoral coordination, interoperability across early warning and decision support systems is crucial to facilitate the exchange of relevant information. 

%In addition, it is equally important to foster better mutual understanding among communities of practice of domain and policy experts who use those systems.
However, interoperability cannot be achieved without a shared conceptual understanding among the %diverse
communities of practice involved in crisis management, including domain experts, policy makers, and operational actors. A key challenge lies in the heterogeneity of terminology and their intended meaning and context of usage (different "perspectives") used across sectors and competence levels (e.g., local, national, and international). 
% Furthermore, crisis management practitioners often are resistant to adopting a different standard, as they are bound to legislation, mandates, or consolidated practices, or because of linguistic barriers.
Furthermore, the adoption of different standards by crisis management practitioners is often complicated by their bound to distinct legislation, mandates, consolidated practices, or by linguistic barriers.

Ontologies are suitable to address these challenges, but existing models target specific sectors individually (most notably health), while crisis management is overlooked. Shared terminologies, such as the United Nations in the context of the Sendai Framework for Disaster Risk Reduction (DRR)~\cite{aitsi2015sendai}, are released as glossaries within reports or annexes to legislation, making them subject to ambiguity. For example, in contrast to the United Nations terminology, the definitions provided in the European regulation on serious cross-border threats~\cite{eu2022_cross_border_threats} use the term "threat" interchangeably with "hazard" (the potential source of harm) and "hazardous event" (the actual manifestation of that hazard). 
Another issue is that existing ontologies typically adopt a non-modular design, making them difficult to extend and adapt to new crisis scenarios (cf. Section \ref{sec:relatedwork}). 

% To address these challenges, we propose the European Crisis Management Ontology (ECMO), which aims to integrate and link existing terminologies across domains into a unified reference framework. Such a connecting layer promotes semantic interoperability across the community of practices and the establishment of common ground, while preserving diversity. In addition to defining and mapping concepts to existing terminologies and controlled vocabularies, our approach leverages the expressive power of Web Ontology Language (OWL) to define concepts not only by name but also by their interrelations, thereby reducing ambiguity. 

%The European Crisis Management Ontology (ECMO) is our proposal to address these challenges: a modular ontology for crisis management, grounded in foundational ontology design patterns and aligned with European standards. 
To address these challenges in support of the ECMP initiative, the Joint Research Centre (JRC) of the European Commission was requested to develop a novel European Crisis Management Ontology (ECMO)\footnote{https://doi.org/10.5281/zenodo.20070268}, which we present in this document as a first prototype version. Funded by the European Commission's Health Emergency Preparedness and Response Authority (DG HERA)\footnote{https://health.ec.europa.eu/health-emergency-preparedness-and-response-hera\_en}, the effort has been carried out in conjunction with the ontology experts who are co-authoring this paper, as well as with the EU policy services that are involved in the joint crisis management initiative.

ECMO is conceived as a modular ontology network for crisis management, grounded in foundational ontology design patterns and aligned with a number of prioritised standards (cf. Section \ref{sec:alignment}).  
In contrast to existing crisis management ontologies, its modular design covers multiple aspects of crisis management while enabling flexible reuse and extension to meet application-specific needs, also beyond the European context. 

Beyond addressing good practices in ontology engineering, such as ontology modularisation, reuse, and alignment with standards, ECMO is designed with a strong policy and societal orientation. It aims at enhancing cross-sectoral interoperability by capturing shared concepts and relationships relevant to crisis preparedness, response, and recovery. 
Its scope %deliberately 
spans multiple domains to reflect the interconnected nature of contemporary crises, including but not limited to:
\begin{itemize}
    \item Public health threats/emergencies and health security;
    \item Food safety and food security, animal and plant health;
    \item Weather extremes and climate change-related threats;
    \item Human-induced environmental degradation and disasters;
    \item Critical infrastructure disruption (including from extraterrestrial hazards such as solar flares);
    \item Civil protection and humanitarian aid;
    \item Security, hybrid threats, and cybersecurity;
    \item Nuclear hazards and incidents;
    \item Armed conflicts and geopolitics;
    \item Migration and refugee crises;
    \item Financial shocks and economic crises;
    \item Societal hazards, human rights, and the rule of law.
\end{itemize}

This broad coverage reflects the ontology's intended applicability across heterogeneous domains and 
stakeholders, supporting its reuse beyond the original development context and reinforcing its 
potential impact in real-world crisis management settings.

In addition, ECMO is positioned to support neuro-symbolic approaches (see Section \ref{sec:extraction}). Its explicit and curated semantics are designed to steer the extraction capabilities of Large Language Models (LLMs). Whereas LLMs can be prompted to generate any kind of content, from text to software code and knowledge graphs, ECMO acts as a qualitative, logically verified prior that grounds LLM outputs in accountable vocabularies and formal axioms. This aligns with emerging neuro-symbolic conditioning approaches~\cite{de2025neurosymbolic}, in which formal ontologies serve as forward pass (the contextual "inference" of LLMs) guardrails for knowledge extraction, 
and reasoning tasks, especially in high-stakes settings where semantic ambiguity carries operational risk.

% To ensure the robustness and relevance of the ontological modules developed, it is essential to validate and fine-tune them against both empirical data and the diverse definitions prevalent across the domains of interest. To this end, we have designed a dedicated application leveraging the advanced capabilities of LLMs. This application facilitates the validation process, and enables iterative refinement, ensuring that the ontological modules remain both accurate and adaptable to evolving requirements.

% [PROJECT AND RELEVANCE TO STAKEHOLDERS]

%[STRUCTURE OF THE PAPER]
The paper is organised as follows. Section~\ref{sec:relatedwork} reviews existing crisis management ontology, identifying key limitations that motivate ECMO. Section~\ref{sec:requirements} outlines the requirements and envisaged coverage of the ontology. Section~\ref{sec:ecmocore} presents ECMO-CORE, covering its foundational alignment with DOLCE, the OWL2 punning pattern for hazard and disaster-risk subtypes, and the frame cluster organisation. Section~\ref{sec:ecmonetwork} describes the ontology network structure and its domain-specific extensions, namely for public health. Section~\ref{sec:alignment} details the alignment methodology and mapping to existing standards, including SNOMED CT, ICD-11, %the UNDRR Hazard Information Profiles, 
and the Sendai Framework Terminology. Section~\ref{sec:extraction} demonstrates the resource’s utility for automated extraction from news sources. Finally, Section~\ref{sec:conclusions} concludes with a discussion of the contributions and an outlook on future work.

% \begin{itemize}
%     \item Does the resource break new ground?
%     \item Does the resource fill an important gap?
%     \item How does the resource advance the state of the art?
%     \item Has the resource been compared to other existing resources (if any) of similar scope?
%     \item Is the resource of interest to the semantic web community?
%     \item Is the resource of interest to society in general?
%     \item How will this resource support the adoption of semantic web technologies?
%     \item What impact will this resource have (or currently has)?
%     \item Is there evidence of usage by a wider community beyond the resource creators or their project? Alternatively (for new resources), what is the resource’s potential for being (re)used?
% \end{itemize}

% ---- Related Work ----
\section{Related work}\label{sec:relatedwork}

%Are there any Ontologies that aim to tackle the same problem? How are they inefficient?

%We divided existing ontologies that are relevant to our work into two groups: 
The literature on crisis management ontologies is growing but fragmented, reflecting the diversity of application domains and modelling approaches. Existing work can be broadly categorised into two categories: (i) cross-sectoral ontologies, and (ii) domain-specific ontologies. 

The Emergency Management Ontology (EMO)~\cite{gao2023emo} is one of the most prominent attempts to provide an ontology for crisis management.
EMO targets the general crisis management lifecycle and its actors. Although EMO is described as a first step into a robust solution, it is limited in some aspects. First of all, the scope of the ontology does not properly account for the case of international organisations that monitor ongoing and potential events to intervene (based on their mandate) when the local capacity is exceeding. Moreover, EMO relies on labour-intensive manual modelling, lacks automated filtering of the massive data streams that responders face, has no fully‑functional prototype for real‑time use, requires constant updates during emergencies, and involves complex multi‑method mapping and possible extra ontologies to achieve the desired semantic interoperability.
Crisis Response Ontology (CROnto)~\cite{bannour2019ontology} focuses on supporting decision-making processes during emergency response operations. It provides a structured representation of crisis situations, actors, and response actions, with particular emphasis on operational coordination. However, its scope remains relatively narrow and primarily oriented towards specific response scenarios. In~\cite{kontopoulos2018ontology}, an ontology for climate crisis management is presented, targeting environmental hazards such as wildfires and extreme weather events. This approach does not generalise easily to other types of crises, such as health emergencies or security threats.
Similarly, the work presented in~\cite{babitski2009ontology} proposes an ontology for information integration in disaster management covering the description of damages caused by the disaster and the resources available to organisations that are relevant for each type of damage. Although it addresses important aspects of semantic integration, the approach is %tightly 
coupled to specific use cases.

% In public-health and epidemiological ontologies, we have a more comprehensive representation of ontologies that cover various domains from different perspectives and aim for diverse goals. 
Several ontologies target specific domains, namely public health and epidemiological, relevant to crisis management.
SimPHO ontology~\cite{okhmatovskaia2012simpho} is an ontology developed for simulation modelling of population health. It aims to reuse medical concepts from existing ontologies and to add health-specific classes (e.g., risk factors and health-policy interventions). As main limitations, SimPHO authors identify a lack of domain coverage and formal evolution. The Simple Event Model + Clinical Element Model (SEM-CEM) ontology~\cite{duncan2015building} aims to solve the chronic problem of linking patient‑specific records across health‑care and public‑health systems when demographic identifiers are inconsistent, incomplete, or change over time. The main limitations of SEM-CEM concern the narrow geographical and institutional scope, the use of simulated data, and the limited coverage of life-course events. The Epidemiology Ontology (EPO)~\cite{pesquita2014epidemiology} aims to address, as the authors describe, the previous incompleteness of %purely
biomedical ontologies in covering epidemiologically specific concepts (e.g., exposure ratios and demographic and epidemiological parameters), by modelling missing terms and linking them to established ontologies. Some of the challenges with EPO identified by the authors include incomplete coverage, insufficient granularity, and no formal evaluation of annotation performance.

%- Also, describe domain-specific resources that tackle the problem at a smaller scale (Semantic Resources).
Beyond domain-specific ontologies, the Sendai Framework Terminology on DRR~\cite{undrr_isc_hips_2025} provides a normative vocabulary of 38 terms that serve as a de-facto cross-sectoral reference for DRR policy. However, this terminology is published only as a glossary - not as a formal ontology - and therefore lacks machine-readable axiomatization. 

Overall, the existing ontologies and domain-specific resources are inefficient to deal with a pan-European high-stakes crisis management setting given the fragmented domain coverage. Each resource aims to tackle a single sector, missing the cross-sectoral concepts we aim to unify (e.g., legal frameworks, finance, humanitarian aid, etc.); lack of alignment with European standards (i.e., not mapping clearly to EU-specific normative frameworks); and insufficient provenance, modularity, and scalability (i.e., lack of governance). These shortcomings motivate the creation of ECMO, which deliberately combines explicit, curated OWL semantics, modular cross-sectoral design, and direct alignment with European standards to overcome the inefficiencies observed in the existing landscape.

% ---- Requirements and Design Principles ----
\section{Requirements and coverage}\label{sec:requirements}
The crisis management landscape is characterised by a wide range of specialised and critical information systems, each a valuable source of heterogeneous information. A key requirement for ECMO is therefore to capture the knowledge 
embedded in these systems, which are widely used by European stakeholders across different sectors. 
In this context, the initial version of the ECMO ontology is guided by an analysis of the following platforms that support crisis preparedness, monitoring, and response and will be integrated or connected with the ECMP "system of systems":

% The crisis management landscape is quite with specialised, critical information systems, each a valuable source of information. Our ontology aims to model the knowledge from key resources and platforms used by key European stakeholders:
\begin{itemize}

% \item The European Crisis Management Platform (ECMP) is an initiative by the European Commission’s Secretariat General in the scope of the Preparedness Union Strategy\footnote{https://commission.europa.eu/topics/preparedness\_en}. It is aimed at integrating and harmonising existing early warning systems and other crisis management relevant tools and information systems and ensure cross-sectoral information exchange and coordination during emergencies.
\item The ECMP Core Platform, a suite of tools aimed at providing harmonised functionalities and workflows for incident management.
\item DG HERA's Advanced Technology for Health INtelligence and Action IT system (ATHINA), an information system developed to strengthen Europe's ability to prevent, detect, and rapidly respond to serious cross-border threats to health, particularly focused on Medical Countermeasures (MCMs).

\item The Early Warning Response System (EWRS)\footnote{https://ewrs.ecdc.europa.eu/}, a secure web-based platform that ensures rapid information exchange and coordinated risk management among EU Member States,
%EWRS is intricately linked with EpiPulse Events, enabling
enabling real-time sharing of situational awareness to support a unified response to serious cross-border health threats as defined by Regulation (EU) 2022/2371. 
\item The EpiPulse Events, the cornerstone system used by the European Centre for Disease Prevention and Control (ECDC)\footnote{https://www.ecdc.europa.eu/en} for event-based surveillance. Strongly linked with EWRS, it allows public health authorities to report and discuss infectious disease incidents, thereby improving threat detection and response strategies.
% \item In the realm of infectious disease surveillance and response, the European Centre for Disease Prevention and Control (ECDC)\footnote{https://www.ecdc.europa.eu/en} utilizes a suite of platforms to enhance situational awareness and facilitate coordinated actions across member states. EpiPulse Events serves as the cornerstone for event-based surveillance, allowing public health authorities to report and discuss infectious disease incidents, thereby improving threat detection and response strategies.
\item The Epidemic Intelligence from Open Sources (EIOS) system\footnote{https://www.who.int/initiatives/eios}, managed by the World Health Organisation (WHO) and developed with the support of the JRC. The platform enables public health intelligence on all hazards according to a One Health approach. Adopted by DG HERA, ECDC and other national and international institutions worldwide, it also allows these organisations to perform collaborative monitoring of public health threats. 
% While DG HERA mainly focuses on potential signals for increased demand of critical medicines and supply chain disruptions, ECDC uses the platform to monitor open sources for infectious disease threats (with plans to integrate it with EpiPulse Events by 2026 to streamline event monitoring), as well to collaborate with other public health agencies at global scale.
% \item Additionally, DG HERA and ECDC and other national and international institutions employ the EIOS system\footnote{https://www.who.int/initiatives/eios}, which is developed in the scope of the Epidemic Intelligence from Open Sources Initiative led by the World Health Organisation (WHO) to support public health intelligence on all hazards according to a One Health approach. While DG HERA mainly focuses on potential signals for increased demand of critical medicines and supply chain disruptions, ECDC uses the platform as monitor open sources for infectious disease threats (with plans to integrate it with EpiPulse Events by 2026 to streamline event monitoring), as well to collaborate with other public health agencies at global scale.

\item The Epi+ platform\footnote{https://www.ecdc.europa.eu/en/publications-data/epi-plus}, designed for European national authorities to facilitate multi-hazard event-based surveillance and communication. The system, which is GDPR-compliant and configurable at the country level, promotes standardisation, collaboration, and interoperability across various levels and sectors and it is intended to be interoperable with EpiPulse Events.
%to enhance EU-level situational awareness.

\end{itemize}

% The review of legislation and science-for-policy recommendations is an essential part of the work, as the systems consuming the and their users must be legally bound to this corpus of prescriptions and underlying knowledge.
% However, the challenge is that this type of source contains a certain degree of semantic ambiguity, in particular used interchangeably to designate both specific instances or manifestations of a given type of entity, and the type of entity itself as a concept. E.g. by distinguishing between “slow onset” and “sudden onset” disasters, the actual intended meaning is that the underlying phenomenon (the hazard) can manifest either suddenly or with a slower progression: this sets expectations on the observed events as they possibly turn into disasters, while not directly describing the feature of the actual disasters.
In addition to those technical systems, the ontology design is informed by the analysis of legislation and science-for-policy recommendations, which define the regulatory and operational context in which crisis management systems are deployed. However, they also introduce challenges due to semantic ambiguity. In particular, terminology is often used interchangeably to refer both to abstract concepts and to their concrete manifestations. For example, distinctions such as "slow-onset" versus "sudden-onset" disasters typically describe the dynamics of underlying hazards rather 
than intrinsic properties of the disaster events themselves.

% As an initial foundation, this paper identifies requirements across the sectors considered, defines the ECMO-CORE Module and some extensions, and provides a first approach to populate the ECMP KG using news sources. 
Based on this analysis, the ontology is designed to meet key requirements. As an initial foundation, this paper introduces the ECMO-CORE module together with selected extensions, and provides a first approach to populating the ECMP knowledge graph using news sources. This demonstrates the feasibility of integrating heterogeneous data sources within a unified semantic framework and highlights the potential of ECMO for supporting real-world crisis management applications. To evaluate the expressiveness and practical relevance of the ontology, a set of competency questions are defined that reflect typical information needed across domains.

Computationally, a competency question  (CQ) is the smallest answerable query that motivates an axiom, a class, or a property in a module. Each CQ has:

\begin{enumerate}
    \item ID - CQ-<MOD>-<P|N><nn> where P is positive (Q+) and N is negative (Q-)
    \item Polarity - Q+ (the ontology should answer YES / return a non-empty result) or Q- (the ontology should answer NO / return empty / a SHACL violation; YES would be a defect)
    \item NL question - natural-language formulation
    \item Exercises - the class/property/axiom or design pattern that the CQ targets
    \item Test kind - SPARQL ASK, SPARQL SELECT, or SHACL shape
\end{enumerate}

CQs have been produced based on the analysis of existing resources (terminologies, glossaries, models), software platforms (e.g. EIOS or DG HERA automated newsfeeds), and information extraction requirements. Each CQ data are encoded in Turtle using a small, pragmatic ontology. Examples of a positive CQ and a negative one are included here:

\begin{verbatim}
cqs:CQ-X-P09 a cqs:CompetencyQuestion ;
    cqs:hasPolarity "positive" ; cqs:targetsModule
        <http://data.europa.eu/h8v/ecmo/core> ;
    cqs:hasNLText "For each governance regime, count 
        disasters in its jurisdiction that exceeded Capacity." ;
    cqs:exercisesPattern "F3×F5×F6 aggregation" ;
    cqs:testKind "sparql_select" ; cqs:expectedAnswer "non_empty" ;
    cpa:hasUnitTest """SELECT ?gov ?j (COUNT(?d) AS ?nExceed) 
    WHERE {
    ?gov a governance:RiskGovernance ;  
        governance:inJurisdiction ?j .
    ?d a disaster:Disaster ; foundation:hasLocation ?j ;
        disaster:exceedsCapacityOf ?cap .
    } GROUP BY ?gov ?j""" .
 
cqs:CQ-X-N01 a cqs:NegativeCompetencyQuestion ;
    cqs:hasPolarity "negative" ; cqs:targetsModule
        <http://data.europa.eu/h8v/ecmo/core> ;
    cqs:hasNLText "Can a Disaster be entailed without
        any Risk it realizesRisk?" ;
    cqs:exercisesPattern "F3×F1 conservativity" ;
        cqs:testKind "sparql_ask" ; cqs:expectedAnswer "false" ;
    cpa:hasUnitTest """ASK { ?d a disaster:Disaster .
        FILTER NOT EXISTS { ?d disaster:realizesRisk 
        ?r . } }""" .
\end{verbatim}

% \begin{enumerate}
% \item Which hazardous events involve the need for addressing the occurrence of both some hydrometeorological hazards and of infectious diseases or pathogens that are food and water-borne? %SPARQL Query + explanation
% \item Which hazardous events caused by a pathogen involve a specific vulnerable population group (e.g. refugees)? %SPARQL Query + explanation
% \item Which events involve the deployment of certain public health and social measures (e.g. contact tracing)? %SPARQL Query + explanation
% \end{enumerate}

% ---- ECMO-CORE ----
\section{ECMO-CORE}\label{sec:ecmocore}

To bridge the different systems, a higher level of abstraction is necessary. 
% However, this is common practice in knowledge engineering and, in fact, recommended to achieve high quality ontology models and interoperability. 
% We rely on foundational ontology design patterns. These are ontology models that address foundational concepts and relations and can be specialised according to the terminological and possibly more restrictive rules that characterise specialised knowledge domains. 
We rely on ontology design patterns that address foundational concepts and relations and can be specialised according to the terminological and possibly more restrictive rules that characterise the addressed knowledge domains.
%Specialising means to align domain specific concepts, for example the concept of Emergency, to the foundational concepts defined by the ontology design patterns.
Following this approach, we designed the ECMO-CORE Ontology, the foundational module of the network of ontological modules, that describes the structure of a crisis pipeline, independent of any specific domain.

\begin{figure}[!h]
    \centering
    \includegraphics[width=0.95\linewidth]{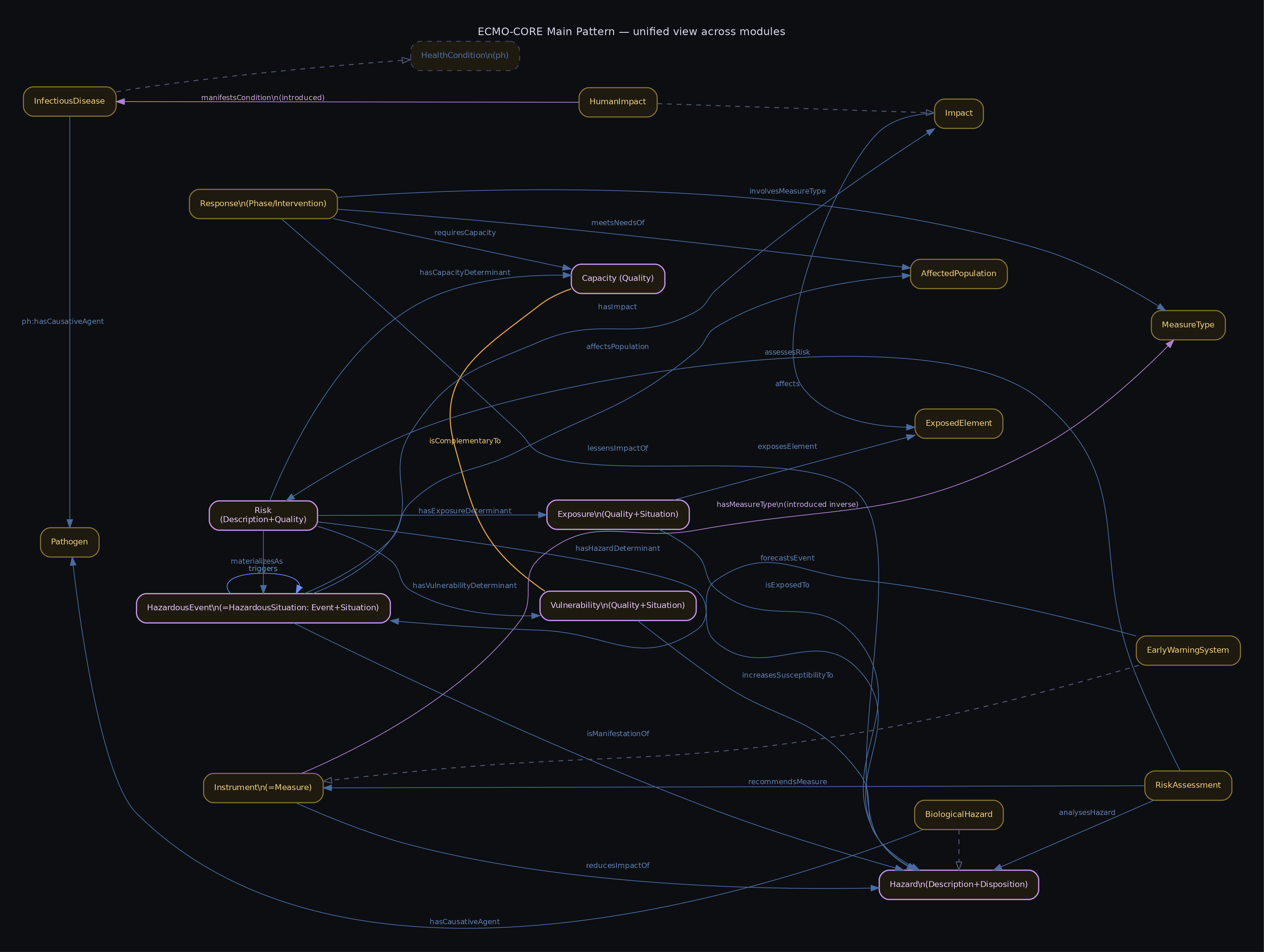}
    \caption{ECMO-CORE: the core conceptual classes and properties.}
    \label{fig:core}
\end{figure}

ECMO-CORE formalises crisis management concepts as OWL2 classes and named individuals, organised into 10 conceptual frames (Risk, Hazard, Disaster, Exposure and Vulnerability, Capacity and Resilience, Governance and Management, Response and Recovery, Risk Reduction Instruments, Underlying Risk Drivers, Impact and Loss), and grounded in the DOLCE foundational ontology via semantic alignments (Figure~\ref{fig:core}). 
%This normative backbone differentiates ECMO from the existing ontologies, which lack alignment with the Sendai reference.

A further methodological contribution of ECMO-CORE concerns the treatment of hazard types and their event-level counterparts. In ontology engineering, domain taxonomies of types (e.g., "Biological Hazard" as a general category) and their spatio-temporal instances (e.g., "the 2026 Chicago meningococcal cluster" as an actual event) pose a well-known modelling tension. ECMO-CORE resolves this by adopting OWL2 punning: each hazard subtype (such as subtypes of \texttt{BiologicalHazard}, \texttt{NaturalHazard}, \texttt{TechnologicalHazard}) is simultaneously a class of situations (providing the OWL-DL type for instances) and a named individual (typed as an instance of an anchor class such as \texttt{BiologicalHazard}, and linked to peers via \texttt{fschema:subsumedUnder}, a transitive property from the Framester schema). This technique – combined with DOLCE's Descriptions and Situations (D\&S) pattern – cleanly separates the intensional (definitional, frame-level) hierarchy from the extensional (instance-level) one, preventing the metaclass conflation that affects many DRR ontologies and enabling sound OWL2 DL reasoning.

% Figure~\ref{fig:core} provides an overview of the conceptual model to represent knowledge of interest for situational awareness and crisis management. It accounts for hazards, hazardous events, exposure, impacts, severity, and measures/interventions. The specific details are discussed in this section. 

% -- Alignement with DOLCE --
\subsection{Foundational alignment with DOLCE design patterns}

ECMO-CORE is aligned with the DOLCE Ultra-Lite (DUL) foundational ontology. The alignment is not merely taxonomic, as ECMO-CORE classes reuse the DUL's design patterns to capture both the descriptive and situational aspects of crisis management concepts. This design choice is particularly relevant for modelling the distinction between general knowledge about hazards and their concrete manifestations. 
As defined by the United Nations Office for DRR (UNDRR), a hazard is "[a type of] process, phenomenon or human activity that may cause loss of life, injury or other health impacts, property damage, social and economic disruption or environmental degradation", while a hazardous event is "the manifestation of a hazard in a particular place during a particular period of time". In ECMO-CORE, the alignment with DUL is operationalised through the following modelling patterns:

\begin{itemize}

%hazard:Hazard
\item \texttt{ecmocore:Hazard} is typed as \texttt{dul:Description} and \texttt{dul:ContextualDisposition} (a contextual attribute of a local situation or event). As a \texttt{Description}, a hazard frames the conditions that hazardous events must satisfy. As a \texttt{ecmocore:ContextualDisposition}, it inheres in the environment as a latent potential for harm. This dual typing operationalises the UNDRR definition, which characterises a hazard as a type of process or phenomenon - i.e. a general description rather than a specific occurrence.

%hazard:HazardousEvent
\item \texttt{ecmocore:HazardousEvent} is a subclass of \texttt{dul:Event} and \texttt{dul:Situation}, carrying existential restrictions on temporal extent (\texttt{dul:TimeInterval}), location (\texttt{dul:Place}), and the hazard it manifests. %(\texttt{ecmocore:isManifestationOf} some Hazard). 
The \texttt{Event} reading captures the occurrence, and the \texttt{Situation} reading captures the fact that certain conditions are simultaneously satisfied.

%expvuln:Exposure
\item \texttt{ecmocore:Exposure} is modelled as a subclass of \texttt{dul:Situation} with restrictions requiring both an exposed element and a hazard, to which the element is exposed (\texttt{ecmocore:isExposedTo} some Hazard).

%response:DisasterRiskReductionPhase
\item \texttt{ecmocore:Mitigation}, \texttt{ecmocore:Prevention}, \texttt{ecmocore:Response}, \texttt{ecmocore:Recovery}, and \texttt{ecmocore:Preparedness} are subclasses of \texttt{ecmocore:DisasterRiskReductionPhase}. Each carries a dual reading: (1) as event-specific actions enacted during a particular hazardous event (Situation/Action reading), and (2) as normative task types prescribed by a \texttt{ecmocore:DisasterRiskManagementPlan} (\texttt{Description} reading, via \texttt{ecmocore:isTaskOf}). This eliminates the need for separate task-type and task-instance class hierarchies.

%response:Intervention
\item \texttt{ecmocore:Intervention} is a subclass of both \texttt{dul:Action} and \texttt{dul:Situation}, with restrictions linking it to a Disaster Risk Management (DRM) phase (\texttt{ecmocore:inDRMPhase}), a capacity (\texttt{ecmocore:involvesCapacity}), and a measure type (\texttt{ecmocore:involvesMeasureType}).

\end{itemize}

% \begin{figure}
%     \centering
%     \includegraphics[width=0.8\linewidth]{Figures/hazard.png}
%     \caption{Modelling of hazards and hazardous events.}
%     \label{fig:hazard}
% \end{figure}

% -- Alignement with DOLCE --
\subsection{OWL2 punning pattern for hazard and disaster-risk subtypes}

Following the classification of hazard information profiles, a number of sub-classes are derived from Hazard, based on UNDRR Hazard Information Profiles (HIPs) taxonomy. The key modelling 
challenge in DRR is that terms like \texttt{ecmocore:BiologicalHazard} function simultaneously as general types (with definitional properties) and as classifiers for specific events. ECMO-CORE addresses this issue through OWL2 punning, where the same IRI serves as both a class and a named individual. This allows the ontology to distinguish, without conflation, between the definitional and factual levels of representation.

At the definitional level, anchor classes such as \texttt{ecmocore:DisasterRisk} and \texttt{ecmocore:Hazard} are modelled as subclasses of \texttt{dul:Description}, capturing general knowledge about types of phenomena. At the factual level, anchor classes such as \texttt{ecmocore:HazardousEvent} and \texttt{ecmocore:Disaster} are modelled as subclasses of \texttt{dul:Situation}, representing concrete occurrences situated in space and time. Within this framework, hazard sub-types are modelled in a dual manner. On the one hand, they are introduced as subclasses of the relevant situation class, enabling standard OWL instance typing for events. On the other hand, the same entities are represented as named individuals of the corresponding anchor description class, thereby participating in an intensional taxonomy. 

These two perspectives are explicitly linked through \texttt{fschema:subsumedUnder}, which encodes specialisation relationships at the level of descriptions. This separation prevents metaclass conflation: \texttt{rdfs:subClassOf} carries the extensional DUL typing for instances, while \texttt{fschema:subsumedUnder} carries the frame-level specialisation hierarchy. The pattern is documented in the ontology's editorial notes and follows the D\&S design pattern from the Ontology Design Patterns community.

% \begin{figure}
%     \centering
%     \includegraphics[width=0.8\linewidth]{Figures/typehazard.png}
%     \caption{Modelling specific types of hazards.}
%     \label{fig:typehazard}
% \end{figure}

% -- Frame Cluster --
\subsection{Frame cluster organisation}

% The 38 Sendai terms and their extensions are organised into 10 named frame clusters (F1–F10), each a \texttt{dul:Description} that groups semantically coherent concepts. Membership is expressed via \texttt{frameCluster} (a sub-property of \texttt{dul:isDescribedBy}). The frame clusters are:
To provide a coherent organisational structure, ECMO-CORE groups the terms into ten named frame clusters (F1–F10). Each frame cluster is modelled as a \texttt{dul:Description}, thereby capturing a shared semantic context that brings together conceptually related terms. Membership of a given term in a frame cluster is expressed through the property \texttt{ecmocore:frameCluster}, defined as a sub-property of \texttt{dul:isDescribedBy}. This frame-based organisation supports both human navigation and systematic coverage verification during ontology engineering.
The resulting frame clusters are presented in Table~\ref{tab:clusters}.

\begin{table}[!t]
    \centering \small
    \caption{Frame clusters description.} \label{tab:clusters}
    \begin{tabular}{l p{2cm} p{6cm} p{4cm}} 
         \textbf{ID} & \textbf{Name} & \textbf{Description} & \textbf{Key concepts} \\ \hline
         F1 & Risk  & Concepts that characterise disaster risk as a relational configuration of hazard, exposure, vulnerability and capacity. & \texttt{DisasterRisk}, \texttt{DisasterRiskAssessment}, \texttt{AcceptableRisk}, \texttt{ResidualRisk} 
         %\texttt{ecmocore:Risk} is the central concept. Sub-types differentiate risk by frequency, scale and residual character after risk reduction measures. 
         \\ \hline
         
         F2 & Hazard & Concepts that describe the physical or anthropogenic processes that may cause loss of life, injury or other health impacts, property damage, social and economic disruption, or environmental damage. & \texttt{Hazard}, \texttt{HazardousEvent}, \texttt{MultiHazardContext}
         %\texttt{ecmocore:Hazard} is the anchor concept. Hazard sub-types are punned as classes plus individuals linked via \texttt{fschema:subsumedUnder}. \texttt{ecmocore:HazardousEvent} is the actual manifestation satisfying the hazard frame. \texttt{ecmocore:MultiHazardContext} groups cases where two or more hazards interact. 
         \\ \hline
         
         F3 & Disaster & Concepts that describe the occurrence, scale and population effects of disasters (serious disruptions of the functioning of a community or society) triggered by hazardous events. Sub-types distinguish onset speed, scale and frequency. & \texttt{Disaster}, \texttt{DisasterManagement}, \texttt{SuddenOnset}, \texttt{SlowOnset}
         %\texttt{ecmocore:Emergency} is an acute episode that may or may not reach disaster severity.  
         \\ \hline
         
         F4 & Exposure and Vulnerability & Concepts that characterise the susceptibility of people, assets and systems to hazard impacts. & \texttt{Exposure}, \texttt{Vulnerability}, \texttt{VulnerabilityDimension}, \texttt{ExposedElement}
         %\texttt{ecmocore:Exposure} describes the presence of people, livelihoods and assets in hazard-prone areas. \texttt{ecmocore:Vulnerability} characterises the conditions that increase susceptibility to harmful consequences. \texttt{ecmocore:VulnerabilityDimension} and its four typed sub-concepts decompose vulnerability into measurable axes. \texttt{ecmocore:ExposedElement} denotes the people, infrastructure or assets in the exposed area.
         \\ \hline
         
         F5 & Capacity and Resilience & Concepts that characterise the ability of communities, organisations and systems to manage disaster risk and adapt to adverse conditions. & \texttt{Capacity}, \texttt{Resilience} 
         %\texttt{ecmocore:Capacity} is the aggregate of all strengths, attributes and resources available to manage and reduce disaster risk. \texttt{ecmocore:Resilience} is the ability to resist, absorb and recover from hazard effects while maintaining essential functions. 
         \\ \hline
         
         F6 & Governance and Management & Concepts that describe the institutional, policy and legal structures through which disaster risk is governed and managed. & \texttt{RiskGovernance}, \texttt{RiskReduction}, \texttt{RiskManagement} %\texttt{ecmocore:RiskGovernance} frames the conditions of institutional coordination. \texttt{ecmocore:RiskReduction} is both the overarching goal frame and the enacted reduction process. \texttt{ecmocore:RiskManagement}, its sub-types, and supporting structures constitute the governance apparatus. 
         \\ \hline
         
         F7 & Response and Recovery & Concepts describing the phases of the DRM cycle from preparedness through response to recovery and reconstruction. & \texttt{Preparedness}, \texttt{Response}, \texttt{Recovery}, \texttt{Reconstruction}, \texttt{Rehabilitation}
         %All phase classes (\texttt{ecmocore:Preparedness}, \texttt{ecmocore:Response}, \texttt{ecmocore:Recovery}, \texttt{ecmocore:Reconstruction}, \texttt{ecmocore:Rehabilitation}) are enacted and constitute an ongoing state. 
         \\ \hline
         
         F8 & Risk Reduction Instruments & Tools, systems and mechanisms used to implement DRR in practice. It includes physical instruments, information and warning systems, financial instruments, and essential services. & \texttt{BuildingCode}, \texttt{EarlyWarningSystem}, \texttt{Retrofitting}, \texttt{RiskTransfer} \\ \hline
         
         F9 & Underlying Risk Drivers & Structural conditions that generate and amplify disaster risk over time. & \texttt{RiskDriver}, \texttt{EnvironmentalDegradation}%\texttt{ecmocore:RiskDriver} is the anchor concept. 
         \\ \hline
         
         F10 & Impact and Loss & Concepts that describe the consequences of disasters on human life, the environment and the economy. & \texttt{Impact}, \texttt{HumanImpact}, \texttt{DisasterDamage}, \texttt{EconomicLoss}
         %\texttt{ecmocore:Impact} is the broadest impact concept. \texttt{ecmocore:HumanImpact} and \texttt{ecmocore:DisasterDamage}, \texttt{ecmocore:EconomicLoss} capture human lost,  physical destruction of built assets and monetary damage, respectively.
         \\ \hline
    \end{tabular}
\end{table}

% -- Observation and measurement pattern --
\subsection{Observation and measurement pattern}

A shared observation infrastructure is implemented with reified observation/estimation patterns. In addition to directly asserting, for instance, that a hazardous event has a certain intensity as a literal value, the ontology provides classes such as \texttt{ecmocore:IntensityObservation}, \texttt{ecmocore:MagnitudeObservation}, and \texttt{ecmocore:SeverityAssessment}, each linked to a \texttt{ecmocore:MeasurementScale} and an \texttt{ecmocore:ObservationSource}. This reification supports provenance tracking, multi-source assessment, and temporal evolution of severity assessments during a crisis.

% ---- ECMO-Extentions ----
\section{ECMO network}\label{sec:ecmonetwork}

% ECMO is a network of interconnected modules designed for clarity, maintainability, and extensibility. The modular architecture enables domain-specific extensions and supports reuse across different sectors.

% -- Public health module --
\subsection{Public health module}

\begin{figure}[!t]
    \centering
    \includegraphics[width=0.95\linewidth]{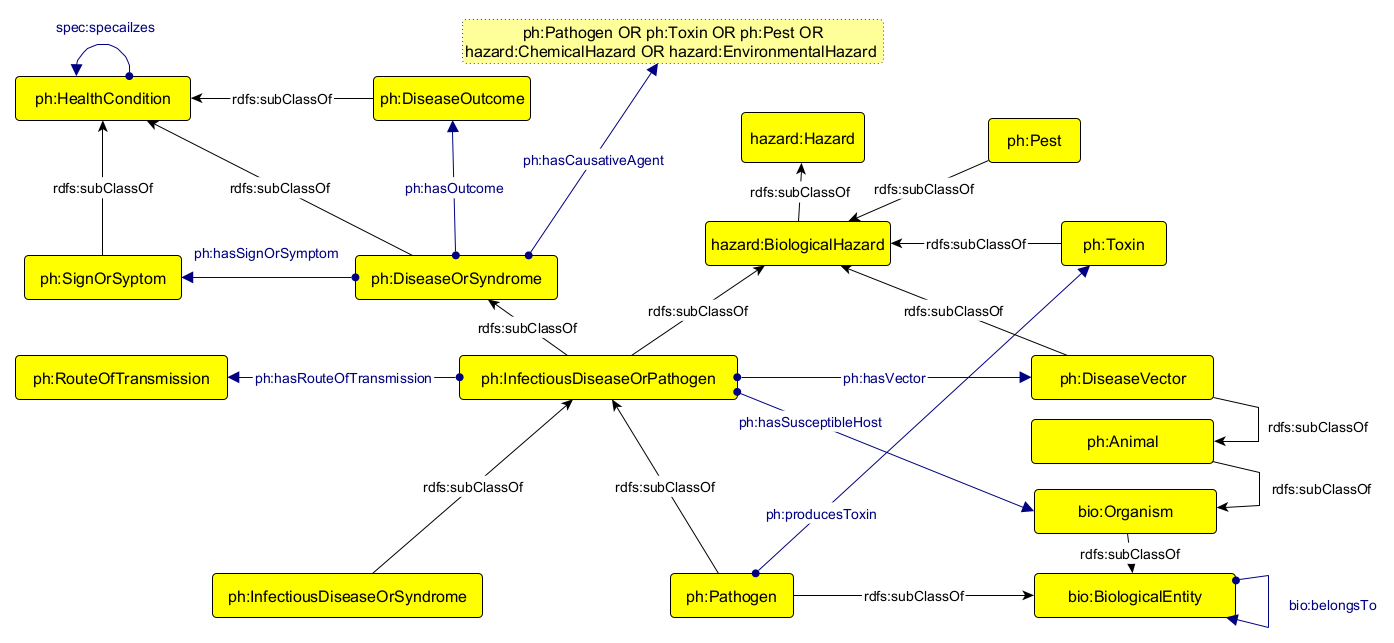}
    \caption{Main classes of the ECMO's public health module.}
    \label{fig:ecmo-ph}
\end{figure}

The public health domain module extends %ECMO-CORE's 
\texttt{ecmocore:BiologicalHazard} with a Pathogen–Disease–Transmission frame (Figure~\ref{fig:ecmo-ph}). Key classes include:

\begin{itemize}
\item \texttt{ph:InfectiousDiseaseOrPathogen} bridges the ECMO-CORE hazard taxonomy with the clinical domain. It is a subclass of \texttt{ph:DiseaseOrSyndrome} and carries a \texttt{dul:hasRole} restriction linking it to \texttt{ecmocore:BiologicalHazard}, so that infectious diseases are simultaneously hazard types (DRR perspective) and clinical entities (health perspective).
\item \texttt{Pathogen} and its sub-types (\texttt{ph:Toxin}, \texttt{ph:Pest}) describe causative agents.
\item \texttt{ph:DiseaseOrSyndrome}, \texttt{ph:InfectiousDiseaseOrSyndrome}, and \texttt{ph:DiseaseOutcome} capture clinical entities.
\item \texttt{ph:RouteOfTransmission}, \texttt{ph:DiseaseVector}, \texttt{ph:SusceptibleHost} capture epidemiological parameters.
\item \texttt{ph:SignOrSymptom} and \texttt{ph:HealthCondition} provide the clinical observation vocabulary.
\item \texttt{bio:Organism} and \texttt{bio:BiologicalEntity} provide basic biological taxonomy support used to classify pathogen vectors and host organisms.
% \item Pathogen risk assessment is supported by \texttt{ecmocore:PathogenCharacteristic} and \texttt{ecmocore:PathogenRiskAssessment}, with sub-classes for \texttt{CaseFatalityRate}, \texttt{Toxicity}, \texttt{Transmissibility}, and related properties.
\end{itemize}

% The bio: namespace provides basic biological taxonomy support (\texttt{bio:Organism}, \texttt{bio:BiologicalEntity}, \texttt{bio:Animal}) used to type pathogen vectors and host organisms, and their specific alignment and extension with clinical ontologies such as SNOMED.

% -- EIOS module --
\subsection{EIOS module}

The EIOS module provides a SKOS-aligned ontology for the EIOS system. At its core, the module defines:

\begin{itemize}
    \item \texttt{eios:Vocabulary} as a subclass of \texttt{skos:ConceptScheme} representing high-level thematic domains (e.g., Human Health, Chemicals and Food Safety).

    \item \texttt{eios:CategoryGrouping} as a subclass of \texttt{skos:Collection} grouping related categories within a vocabulary. Groupings are linked to their members using \texttt{skos:member}.

    \item \texttt{eios:Category} as a subclass of \texttt{skos:Concept}, representing specific entities or phenomena (e.g., Anthrax, Taenia, Adenovirus).
\end{itemize}

In addition to this SKOS layer, the EIOS module also models the event-detection workflow. The class \texttt{eios:OpenSourceData} captures information objects collected from public sources, while \texttt{eios:InformationSource} represents the originating source. Such data may generate a \texttt{eios:Signal}, denoting an anomalous indication requiring further assessment. Signals can \texttt{eios:pertainsToEvent} instances of \texttt{eios:PublicHealthEvent}. 

This formalisation enables not only automated mappings and alignment of EIOS categories to biomedical ontologies, but also semantic enrichment of event-based surveillance data, supporting advanced queries and analytics.

% -- DG HERA module --
\subsection{DG HERA module}

The DG HERA module in ECMO provides a model for evaluating medical countermeasures and pathogen risk in the context of public health emergencies. The key features of this module include:

\begin{itemize}
    \item \texttt{hera:PathogenRiskAssessment} is the core class and represents the risk assessment of a pathogen according to specific criteria. Subclasses define the main assessment dimensions, namely \texttt{hera:PHEICPotential},  \texttt{hera:Transmissibility}, \texttt{hera:PopulationImmunity}, \texttt{hera:Pathogenicity}, \texttt{hera:AvailabilityOfVaccines}, \texttt{hera:AvailabilityOfTherapeutics}, and \texttt{hera:EUCrossBorderThreatPotential}.
    
    \item Each assessment dimension is linked to a \texttt{hera:AssessmentCriterion} (e.g., \texttt{hera:PHEICCriterion}).

    \item For each dimension, it is defined a scale class (e.g., \texttt{hera:PHEICPotentialScale}) and a set of ordered qualitative values (e.g., \texttt{hera:Low\_PHEICPotential}).

    \item Object properties such as  \texttt{hera:hasPHEICPotential}, to link a \texttt{hera:Pathogen} to its risk assessment values.
    
\end{itemize}

The DG HERA module enables the prioritisation of health threats based on multiple risk dimensions, supporting decision-making for medical countermeasures deployment.

% ---- Identification and Mapping of Existing Standards ----
%\section{Identification and mapping of existing standards}
\section{Alignment with existing standards}\label{sec:alignment}

ECMO adopts a semi-automatic mapping strategy aimed at facilitating the integration of external standards and sources of contextual information for cross-domain reasoning. The adopted methodology follows a multi-step workflow, where automated ontology alignment tools generate candidate correspondences, which are subsequently curated and validated by domain experts. 
%ECMO adopts a semi-automatic mapping strategy, in which automated ontology alignment tools generate candidate correspondences, which are subsequently curated and validated by domain experts. 
% The adopted methodology follows a multi-step workflow. 
The process begins with the selection of ontologies and terminological resources to be aligned. This step considers domain relevance and conceptual coverage.
Following ontology selection, automated ontology matching systems are employed to generate initial candidate correspondences between concepts across the selected standards. 
The candidate mappings produced by different systems are subsequently aggregated. The use of multiple alignment tools increases mapping coverage and reduces reliance on a single matching system, allowing complementary matching techniques to capture a broader range of potential semantic relations. During this step, mappings with higher confidence scores or stronger agreement among systems are prioritised.

The resulting candidate alignments are then subjected to expert %curation and 
validation. Expert assessment plays a critical role in the process, as incorrect mappings may lead to inappropriate operational decisions in crisis scenarios. Validated mappings can be represented using standardised semantic web mechanisms such as \texttt{owl:equivalentClass}, \texttt{owl:sameAs} and \texttt{rdfs:subClassOf}, as well as SKOS relations including \texttt{skos:exactMatch}, \texttt{skos:closeMatch}, and \texttt{skos:relatedMatch}.
Finally, logical consistency verification is performed using reasoning tools to ensure that the identified mappings do not introduce contradictions into the %ECMO 
ontology.

% -- Linking to ECMO-PH --
\subsection{HIPS, SNOMED, ICD-11}

% As a first application domain, public health was selected to pilot the identification and mapping of relevant standards. 
Three resources are identified as particularly relevant in the public health domain: (i) HIPs~\cite{undrr_isc_hips_2025} developed to support the implementation of DRR and risk-informed investment, aligned with the Sendai Framework for DRR 2015–2030; (ii) SNOMED CT~\cite{benson2012principles} as a comprehensive clinical terminology; (iii) ICD-11~\cite{harrison2021icd} as an international standard for disease classification. Together, these resources cover complementary perspectives on hazards and health conditions.

The public health module models pathogens and diseases as individuals, each annotated with preferred labels and biological taxonomy, and links them to SNOMED CT, EIOS, and HIPS. The mappings between these resources were obtained by applying LogMap~\cite{jimenez2011logmap}. 
The following turtle snippet illustrates this approach with the case of \texttt{Hepatitis B}, represented as an ECMO individual and aligned with corresponding entities in ICD-11, SNOMED CT, and HIPS through \texttt{owl:sameAs} triples:

\begin{lstlisting}[label={lst:hepb-mapping}]
@prefix ph: <http://data.europa.eu/h8v/ecmo/public-health/> .
@prefix hips: <https://www.undrr.org/terms/hips/> .
@prefix icd11: <http://id.who.int/icd/entity/> .
@prefix snomed: <http://snomed.info/id/> .
@prefix owl: <http://www.w3.org/2002/07/owl#> .
@prefix skos: <http://www.w3.org/2004/02/skos/core#> .
@prefix spec: <http://www.ontologydesignpatterns.org/cp/owl/specialization.owl#> .
ph:HepatitisB a owl:NamedIndividual , 
               ph:InfectiousDiseaseOrPathogen ;
    ph:hasSusceptibleHost ph:Humans ;
    spec:specializes ph:Hepatitis ;
    owl:sameAs icd11:352087872,
        snomed:66071002,
        hips:bi0212 ;
    skos:prefLabel "Hepatitis B"@en .
\end{lstlisting}

% -- Linking ECMO-CORE to the Sendai Framework Terminology--
\subsection{Sendai Framework Terminology}

ECMO-CORE also contains the full set of 38 Sendai DRR terms, each annotated with \texttt{skos:definition} carrying the official UNDRR definition, and \texttt{rdfs:label} in multiple languages. The \texttt{dcterms:source} property at the ontology level traces back to the UNDRR DRR Glossary. This direct axiomatisation complements the external SKOS based mappings to SNOMED CT, ICD-11 and EIOS categories, providing a normative anchor that the mapped resources can reference.

\section{Automated extraction from news sources} \label{sec:extraction}

A key functionality of ECMO lies in its role as a prior for knowledge graph extraction from an actual EIOS news item. E.g., a source article reports that the Chicago Department of Public Health (CDPH) identified four cases of Neisseria meningitidis infections in adults since January 15, 2026, of which two were among people living in a temporary homeless shelter. All patients developed septicemia, and one additionally had meningitis.

% The extraction follows a neurosymbolic pipeline that combines automated frame-semantic parsing with ECMO-guided domain alignment and expert validation.
First, the text is parsed by Text2AMR2FRED~\cite{text2amr2fred} to produce an initial graph of entities, events, and role structures. The base graph is then subjected to ontological verification and domain alignment against the ECMO ontology network using a set of systematic heuristics from the POLANYI++ methodology~\cite{de2025neurosymbolic}. 

The article itself is represented as \texttt{eios:OpenSourceData}, generating an \texttt{eios:Signal} for a \texttt{eios:PublicHealthEvent}, thereby linking the extracted knowledge to the EIOS vocabulary.
As an example of non-trivial extraction, the ontology-augmented pipeline resolves the frequent conflation between pathogen and disease by distinguishing N.~meningitidis as a \texttt{ph:Pathogen} and Invasive Meningococcal Disease as a \texttt{ph:InfectiousDiseaseOrPathogen}. 
More details are in the GitHub\footnote{https://github.com/aldogangemi/ecmo}.%supplementary material.
\section{Conclusions and future work}\label{sec:conclusions}

This paper presents ECMO, a modular OWL2-based ontology network designed to support semantic interoperability across heterogeneous crisis management information systems. Built on foundational ontology design patterns and aligned with DOLCE, ECMO-CORE provides a cross-sectoral reference model. Its frame clusters organise core crisis management concepts, from hazard and exposure to governance, resilience, and impact. The public health extension, the EIOS module, and the DG HERA module demonstrate the practical extensibility of the modular architecture, while the semi-automatic alignment methodology enables scalable integration with SNOMED CT, ICD-11, and the UNDRR HIPs. The end-to-end extraction pipeline, illustrated through a real EIOS surveillance article, provides concrete evidence of the ontology's utility as a guardrail for knowledge extraction.
ECMO is a relevant resource both to the Semantic Web community and to European crisis preparedness policy.

% Future work will include extension to other domains and mapping to other existing standards when available, the establishment of architectures and governance approaches for continuous knowledge extraction, update and curation, with link to intelligence gathering, and the development of a graphical interfaces for visualisation. 
Future work encompasses two main directions. First, an LLM-assisted validation workflow will support iterative and collaborative expert refinement of ontological definitions. Experts will evaluate definitions in their domain, submit alternatives where needed, and the resulting inputs will be aggregated to identify convergence and divergence across domains, with LLMs synthesising consolidated definitions that resolve detected ambiguities. 
Second, an end-to-end ECMO-compliant knowledge graph pipeline will be deployed, covering ingestion from sources such as EIOS and GDACS, preprocessing and harmonisation, triplification via RML mappings into RDF, SHACL-based validation, and publication through SPARQL endpoints and RESTful APIs.

% The presented work represents a first foundation for facilitating semantic interoperability across crisis management information systems in Europe and beyond. It also paves the road a wide range of research activities aimed
% \begin{itemize}
%     \item Extension to other domains and mapping to other existing standards when available
%     \item Establishment of architectures and governance approaches for continuous knowledge extraction, update and curation, with link to intelligence gathering
%     \item User experience aspects: how to present ontology and knowledge graph to support sense making and advanced business analysis.
%     \item Agentic AI consumption and discovery of anomalies and patterns.
% \end{itemize}

\section*{Declaration of use of Generative AI}
During the preparation of this work, the author(s) used ChatGPT and Grammarly for grammar and spelling check, and for paraphrasing and rewording.

\bibliographystyle{splncs04}
\bibliography{mybibliography}

\end{document}